\pdfoutput=1
\documentclass{easychair}

\newcommand{\xf}[1]{Figure~\ref{#1}}

\newcommand{\flucid}{{Forensic Lucid\index{Forensic Lucid}}}

\newcommand{\java}{{Java\index{Java}}}

\newcommand{\file}[1]{\url{#1}\index{Files!#1}}
\newcommand{\tool}[1]{\texttt{#1}\index{Tools!#1}}

\newcommand{\api}[1]{\texttt{#1}\index{API!#1}}

\newcommand{\marf}[0]{MARF\index{MARF}\index{Frameworks!MARF}\index{Libraries!MARF}}

\newcommand{\lucidL}[1]{{$\mathit{Lucid}$}($L$) }

		{}

\def\myvert{\raise 2.27pt \hbox{\vrule depth 0pt height 8pt width 0.2mm}}
\def\myarrow{\hspace*{0.43mm}%
             \raise 2.29pt\hbox{\vrule depth 0pt height 8pt width 0.16mm}%
             \hspace*{-0.32mm}%
             $\longrightarrow$
             \ %
             }

\begin{document}

\title{Writer Identification Using Inexpensive Signal Processing Techniques}
\titlerunning{Writer Identification Using Inexpensive Signal Processing Techniques}

\author{
Serguei A. Mokhov\\
Computer Science\\and Software Engineering,\\
Concordia University, Montreal, Canada,\\
Email: \url{mokhov@cse.concordia.ca}\\
\and
Miao Song\\
Graduate School,\\
Concordia University,\\Montreal, Canada,\\
Email: \url{m_song@cse.concordia.ca}\\
\and
Ching Y. Suen\\
Centre for Pattern Recognition\\and Machine Intelligence,\\
Concordia University, Montreal, Canada,\\
Email: \url{suen@cenparmi.concordia.ca}
}

\authorrunning{Mokhov, Song and Suen}

\maketitle

\begin{abstract}
We propose to use novel and classical audio and text signal-processing and otherwise techniques
for ``inexpensive'' fast writer identification tasks of scanned hand-written documents ``visually''.
The ``inexpensive'' refers to the efficiency
of the identification process in terms of CPU cycles while preserving decent accuracy for
preliminary identification. This is a comparative study of multiple algorithm combinations
in a pattern recognition pipeline implemented in {\java} around an open-source Modular Audio Recognition Framework ({\marf})
that can do a lot more beyond audio. We present our preliminary experimental findings in such an identification task.
We simulate ``visual'' identification by ``looking'' at the hand-written document as a whole
rather than trying to extract fine-grained features out of it prior classification.
\end{abstract}

\section{Introduction}
\label{sec:introduction}

\subsection{Problem Statement}

Current techniques for writer identification often rely on the classical
tools, methodologies, and algorithms in handwriting recognition (and in
general in any image-based pattern recognition) such as skeletonizing,
contouring, line-based and angle-based feature extraction, and many others.
Then those techniques compare the features to the ``style'' of features a given writer may have
in the trained database of known writers. These classical techniques are,
while highly accurate, are also time consuming for bulk processing of a
large volume of digital data of handwritten material for its preliminary
or secondary identification of who may have written what.

\subsection{Proposed Solution}

We simulate ``quick visual identification'' of the hand-writing of the writer
by looking at a page of hand-written text as a whole
to speed up the process of identification, especially
when one needs to do a quick preliminary classification of a large volume
of documents. For that we treat the sample pages as either 1D or 2D
arrays of data and apply 1D or 2D loading using various loading
methods, then 1D or 2D filtering, then in the case of 2D filtering,
we flatten a 2D array into 1D prior feature extraction, and then
we continue the classical feature extraction, training and classification
tasks using a comprehensive algorithm set within Modular Audio
Recognition Framework ({\marf})'s implementation, by roughly treating
each hand-written image sample as a wave form as in e.g. in speaker 
identification. We insist on 1D as it is the baseline storage
mechanism for {\marf} and it is less storage consuming while sufficient
to achieve high accuracy in the writer identification task.

This approach is in a way similar to the one where {\marf} was applied
to file type analysis for forensic purposes~\cite{marf-file-type} using machine learning
and assuming each file is a sort of a signal on Unix systems as compared to the traditional
\tool{file} utility~\cite{file-man,file-site}.

\subsection{Introduction to MARF}

Modular Audio Recognition Framework ({\marf}) is an open-source collection of pattern recognition APIs and their implementation for
unsupervised and supervised machine learning and classification written in {\java}~\cite{marf02,marf,marf-ccct08,marf-cisse07,marf-ai08,marf-c3s2e08}.
One of its design purposes is to act as a testbed to try out
common and novel algorithms found in literature and industry for sample loading, preprocessing, feature extraction,
and training and classification tasks. One of the main goals and design approaches
of {\marf} is to provide scientists with a tool for comparison of the algorithms in a homogeneous environment and
allowing the dynamic module selection (from the implemented modules) based on the configuration options supplied by applications.
Over the course of several years {\marf} accumulated a fair number of implementations for each of the pipeline
stages allowing reasonably comprehensive comparative studies of the algorithms {\em combinations},
and studying their {\em combined} behavior and other properties when used for various pattern recognition tasks.
{\marf} is also designed to be very configurable while keeping the generality and some sane default settings
to ``run-off-the-shelf'' well.
{\marf} and its derivatives, and applications were also used beyond audio
processing tasks due to the generality of the design and
implementation in~\cite{dmarf06,dmarfsnmp07,marf-gipsy-distributed-ispdc08,jdsf-wmsci08,sqlrand-key-management-sac09}
and other unpublished or in-progress works.

\begin{figure*}[t!]
	\centering
	\includegraphics[width=\textwidth]{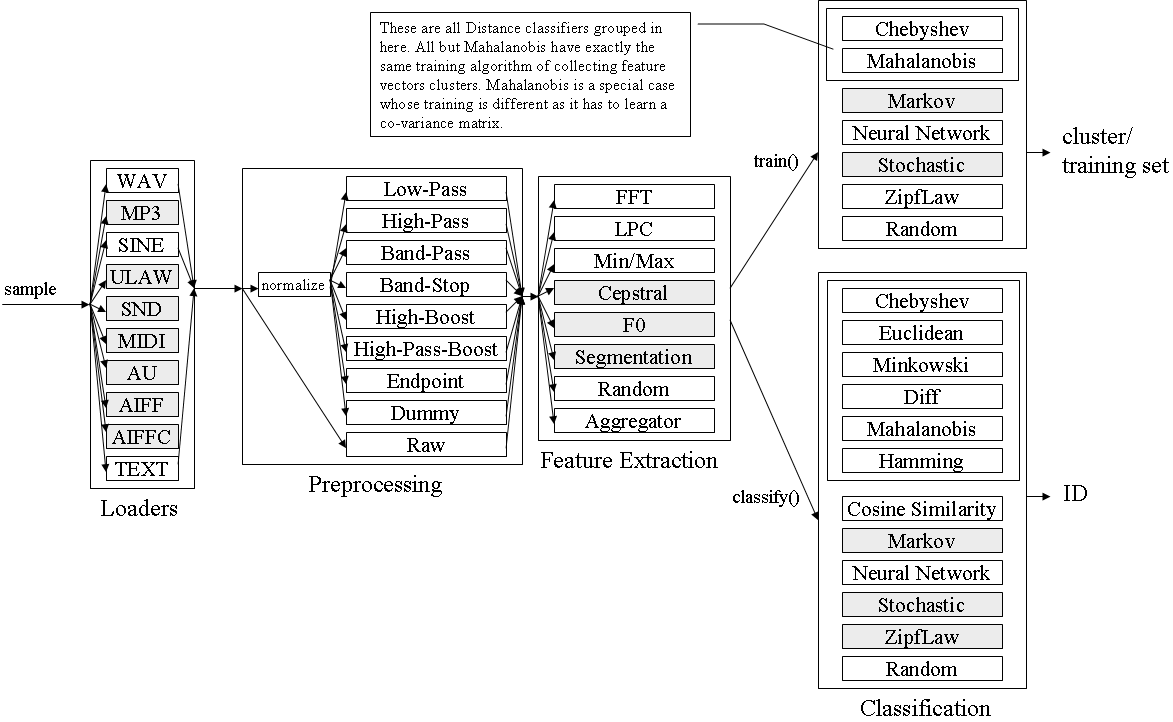}
	\caption{MARF's Pattern Recognition Pipeline}
	\label{fig:pipeline-flow}
\end{figure*}

\subsubsection{Classical Pattern Recognition Pipeline}

The conceptual pattern recognition pipeline shown in \xf{fig:pipeline-flow} depicts
the core of the data flow and transformation between the stages of the {\marf}'s pipeline.
The inner boxes represent most of the available concrete module implementations or stubs.
The grayed-out boxes are either the stubs or partly implemented.
The white boxes signify implemented algorithms.
Generally, the whole pattern recognition process starts by loading a sample (e.g. an
audio recording in a wave form, a text, or image file), preprocessing it (removing noisy and ``silent'' data
and other unwanted elements), then extracting the most prominent features, and finally either
training of the system such that the system
either learns a new set of a features of a given subject or actually classify
and identify what/how the subject is.
The outcome of training is either a collection of some form of feature vectors or their
mean or median clusters, which a stored per subject learned. The outcome of classification
is a 32-bit unique integer usually indicating who/what the subject the system believes is.
{\marf} designed to be a standalone \file{marf.jar} file required to be usable
and has no dependencies on other libraries. Optionally, there is
a dependency for debug versions of \file{marf.jar} when JUnit~\cite{junit} is used
for unit testing.

\subsubsection{Algorithms}
\label{sect:algos}

{\marf} has actual implementations of the framework's API
in a number of algorithms to demonstrate its abilities in various pipeline stages and modules.
There are a number of modules that are under the process of implementation or porting from
other project for comparative studies that did not make it to this work at the time of its writing.
Thus, the below is an incomplete summary of implemented algorithms corresponding to the \xf{fig:pipeline-flow}
with a very brief description:

\begin{itemize}
\item
Fast Fourier transform (FFT), used in FFT-based filtering as
well as feature extraction~\cite{dspdimension}.
\item
Linear predictive coding (LPC) used in feature extraction.
\item
Artificial neural network (classification).
\item
Various distance classifiers (Chebyshev, Euclidean, Minkowski~\cite{abdi-distance}, Mahalanobis~\cite{mahalanobis-distance},
Diff (internally developed within the project, roughly similar in behavior to the UNIX/Linux \tool{diff} utility~\cite{diff}),
and Hamming~\cite{hamming-distance}).
\item
Cosine similarity measure~\cite{cosine-similarity-tutorial,cosine-similarity-euclidean-distance},
which was thoroughly discussed in~\cite{khalifemcthesis04} and often produces the best accuracy in
this work in many configurations (see further).
\item
Zipf's Law-based classifier~\cite{zipfslaw}.
\item
General probability classifier.
\item
Continuous Fraction Expansion (CFE)-based filters~\cite{shivaniharidasbhat06}.
\item
A number of math-related tools, for matrix and vector processing,
including complex numbers matrix and vector operations, and statistical estimators
used in smoothing of sparse matrices (e.g. in probabilistic matrices or Mahalanobis
distance's covariance matrix). All these are needed for {\marf} to be self-contained.
\end{itemize}

\section{Methodology}

To enable the experiments in this work and their results we required
to do the alteration of the {\marf}'s pipeline through its plug-in
architecture. We outline the modifications and the experiments
and conducted using a variety of options.

\subsection{Modified MARF's Pipeline}

There are slight modifications to the pipeline that were required
to {\marf}'s original pipeline in order to enable some of the experiments
outlined below for the writer identification tasks. Luckily, due to
{\marf}'s extensible architecture we can do those modifications as
plug-ins, primarily for sample loading and preprocessing, that we
plan on integrating into the core of {\marf}.

\subsubsection{Loaders}

We experiment with a diverse scanned image sample loading mechanisms to
see which contribute more to the better accuracy results and the most efficient.
There is a naive and less naive approach to do so. We can treat the
incoming sample as:

\begin{itemize}
\item an image, essentially a 2D array, naturally
\item a byte stream, i.e. just a 1D array of raw bytes
\item a text file, treat the incoming bytes as text, also 1D
\item a wave form, as if it is encoded WAVE file, also 1D
\end{itemize}

Internally, regardless the initial interpretation of the scanned
hand-written image samples, the data is always treated as some
wave form or another. The initial loading affects the outcome
significantly, and we tried to experiment which one yields
better results, which we present as options. For this to
work we had to design and implement \api{ImageSampleLoader}
as an external to {\marf} plug-in to properly decode
the image data as image and return a 2D array representation
of it (it is later converted to 1D for further processing).
We adapt the \api{ImageSample} and \api{ImageSampleLoader} previously
designed for the \api{TestFilters} application of {\marf} for 2D filter
tests.
The other loaders were already
available in {\marf}'s implementation, but had to be subclassed or wrapped around
to override some settings. Specifically, we have
a \api{ByteArrayFileReader}, \api{TextLoader}, and \api{WAVLoader} in the core {\marf}
that we rely upon.
For the former, since it does not directly implement the \api{ISampleLoader}
interface, we also create an external wrapper plug-in, \api{RawSampleLoader}.
The \api{TextLoader} provides options for loading the data
as uni-, bi-, and tri-gram models, i.e. one sample point
consists of one, two, or three characters. The \api{WAVLoader}
allows treating the incoming sample at different sample
rates as well, e.g. 8000 kHz, 16000 kHz, and so on. We
had to create a \api{TIFFtoWAVLoader} plug-in for this research
work to allow the treatment of the TIFF files as WAV with
the proper format settings.

\subsubsection{Filters}

The Filter Framework of {\marf} and its API represented by the
\api{IFilter} interface has to be invoked with the 2D versions
of the filters instead of 1D, which is a sufficient default
for audio signal processing. The Filter framework has a 2D
API processing that can be applied to images, ``line-by-line''.
The 2D API of \api{IFilter} returns a 2D results. In order for
it to be usable by the rest of the pipeline, it has to be
``flattened'' into a 1D array. The ``flattening'' can be done
row-by-row or column-by-column; we experiment with both ways
of doing it. Once flattened, the rest of the pipeline process
functions as normal. Since there is no such a default preprocessing
module in the core {\marf}, we implement it as a preprocessing
plug-in in this work, which we call \api{Filter2Dto1D}. This
class implements \api{preprocess()} method to behave the
described way. This class in itself actually does not do much,
but instead the FFT-based set of filters is mirrored from the
core {\marf} to this plug-in to adhere to this new implementation
of \api{preprocess()} and at the same time to delegate all the
work to the core modules. Thus, we have the base
\api{FFTFilter2Dto1D}, and the concrete \api{LowPass2Dto1D},
\api{HighPass2Dto1D}, \api{BandStop2Dto1D}, and \api{BandPass2Dto1D}
FFT-based filters.
The CFE-based filters require further testing at this point
and as such were not included in the experiments.

\subsubsection{Noise Removal}

We employ two basic methodologies of noise removal in our experiments:
(1) we either remove the
noise by loading the ``noise'' sample, a scanned ``blank'' sheet
with no writings on it. Subtracting the frequencies of this
noise sample from the incoming samples gives us the net effect
of large noise removal. This FFT sample-based noise remover is only
effective for the 2D preprocessing operations.
Implementation-wise, we implement it in the \api{SampleBasedNoiseRemover}
preprocessing plug-in class.
(2) We compare that to the default noise removal in {\marf} that
is constructed by application of the plain 1D low-pass FFT filter.

\subsection{\api{WriterIdentApp}}

We provide a testing application, called \api{WriterIdentApp}, to do 
all the experiments in this work and statistics gathering.
The application is a writer-identification-oriented fork of \api{SpeakerIdentApp}
present within {\marf}'s repository~\cite{marf} for speaker
identification task. The application has been amended
with the options to accept the four loader types instead of two,
noise removal by subtraction of the noise sample, and the 2D
filtering plug-ins. The rest of the application options are
roughly the same as that of \api{SpeakerIdentApp}. Like all
of {\marf} and its applications, \api{WriterIdentApp} will
be released as open-source and can be made available to the
willing upon request prior that.

\subsection{Resolution}

We vary the resolution of our samples as 600dpi,
300dpi, and 96dpi in our experiments to see how it
affects the accuracy of the identification. The
samples are both grayscale and black-and-white.

\section{Testing, Experiments, and Results}

The handwritten training samples included two pages
scanned from students' quizzes. The testing performed
on the another, third page of the same exam for each
student. The total number of students's exams in class
studied is 25.

\subsection{Setup}

\noindent
In the setup we are testing multiple permutation of configurable
parameters, which are outlined below.

\subsubsection{Samples}

\noindent
The samples are scanned pages letter-sized as uncompressed TIFF images of the following resolutions and color schemes:

\begin{itemize}
\item 600 dpi grayscale, black-and-white
\item 300 dpi grayscale, black-and-white
\item 96 dpi grayscale, black-and-white
\end{itemize}

\subsubsection{Sample Loaders}

\begin{itemize}
\item Text loader: unigram, bigram, trigram
\item WAVE loader: PCM, 8000 kHz, mono, 2 bytes per amplitude sample point
\item Raw loader: byte loader (1-byte, 2-byte, 4-byte per sample point)
\item TIFF Image 2D loader
\end{itemize}

Byte loader and text loader are similar but not identical.
In {\java} characters are in UNICODE and occupy physically two bytes
and we use a character-oriented reader to do so.
In the byte loader, we deal with the raw bytes and our ``ngrams''
correspond to the powers of 2.

\subsubsection{Preprocessing}

\begin{itemize}
\item 1D filtering works with 1D loaders, and low-pass FFT filter acts as a noise remover
\item 2D filtering covered by a plug-in with 2D FFT filters and noise sample subtraction
\item Flattening of the 2D data to 1D by row or column
\end{itemize}

\subsubsection{Feature Extraction and Classification}

\noindent
The principle fastest players in the experimentation so-far
were primarily the distance and similarity measure classifiers
and for feature extraction FFT, LPC and min/max amplitudes.
All these modules are at their defaults as defined by
{\marf}~\cite{marf,marf-cisse07,marf-ccct08,marf-ai08,marf-c3s2e08}.

\subsection{Analysis}

Generally, it appears the 2D versions of the combinations produce higher
accuracy. The text-based and byte-based loaders perform at the average
level and the wave form loading slightly better. The black-and-white
images at all resolutions obviously load faster as they are much
smaller in size, and even the 96 dpi-based image performed very well
suggesting the samples need not be of the highest quality. Algorithm
combinations that had silence removed after either 1D or 2D based 
noise removal contributed to the best results by eliminating ``silence
gaps'' (in the image strings of zeros, similar to compression) thereby
making samples looking quite distinct. The noise-sample based removal,
even eliminates the printed text and lines of the handed-out exam
sheets keeping only hand-written text and combined with the silence
removal pushes the accuracy percentage even higher.

The experiments are running on two major hardware pieces: a Dell
desktop and a server with two 2 CPUs.
Run-time that it takes to train the system on 50 600dpi grayscale
samples (35Mb each) is varied between 15 to 20 minutes on a Dell
Precision 370 workstation with 1GB of RAM and 1.5GHz Intel Pentium 4
processor running Fedora Core 4 Linux. For the testing samples, it
takes between 4 and 7 seconds depending on the algorithm combination
for the 35Mb testing samples.	All the sample files were read off
a DVD disk, so the performance was less optimal than from a hard disk.
In the case of the server with two Intel Pentium 4 CPUs, 4GB of ram and the four
processing running it takes 2-3 times faster for the same amount.
96dpi b/w images take very fast to process and offer the best response
times.

\section{Conclusion}

As of this writing due two numerous exhaustive combinations and about 600
runs per loader most of the experiments and some testing are still underway
and are expected to complete within a week.
The pages with the list of resulting tables
are obviously not fitting within a 6-page conference paper, but will be made
available in full upon request. Some of the fastest results have come
back entirely, but for now they show disappointing accuracy 
performance of 20\% correctly identified writiers in our settings, which
is way below expected from our hypothesis.
Since the results are incomplete, the authors are reviewing them as
they come in and seek the faults in the implementation and data.
The complete set of positive or negative outcomes will be summarized
in the final version of the article.

\subsection{Applications}

\noindent
We outline possible applications of our quick classification
approach:

\begin{itemize}
\item
Students' exams verification in case of fraud claims
to quickly sort out the pages into appropriate bins

\item
For large amount of scanned checks (e.g. same as banks
make available on-line).
Personal checks identification can be used to see
if the system can tell they are written by
the same person. In this case the author used
his personal check scans due to their handy
availability.

\item
Quick sorting out of hand-written mail.

\item
Blackmail investigation when checking whether some
letters with threats were written by the same person
or who that person might be by taking sample handwriting
samples of the suspects in custody or investigation.

\end{itemize}

\subsection{Future Work}

\begin{itemize}
\item
Further improve recognition accuracy by investigating more algorithms and their properties.
\item
Experiment with the CFE-based filters.
\item
Automation of training process for the sorting purposes.
\item
Export results in {\flucid} for forensic analysis.
\end{itemize}

\subsection{Improving Identification Accuracy}

So far, we did a quick way of doing writer authentication
without using any common advanced or otherwise image processing 
techniques, such as contouring, skeletonizing, etc. and the related
feature extraction, such as angles, lines, direction, relative position
of them, etc. We can ``inject'' those approaches into the available
pipeline if we can live with slower processing speeds due to the
additional overhead induced by the algorithms, but improve the
accuracy of the identification.

\section*{Acknowledgments}

This work is partially funded by NSERC, FQRSC, and Graduate School and the Faculty of Engineering
and Computer Science, Concordia University, Montreal, Canada.

\bibliographystyle{plain}
\bibliography{marf-writer-ident-arXiv}

\end{document}